\documentclass[11pt,a4paper]{article}
\usepackage[hyperref]{emnlp2020}
\usepackage{times}
\usepackage{latexsym}

\usepackage{microtype}
\usepackage{graphicx}
\usepackage{subfigure}
\usepackage{booktabs} 
\usepackage{amsmath}
\usepackage{multirow}
\usepackage{enumitem} 
\usepackage{bbm}
\usepackage{amssymb}
\usepackage{pifont}
\usepackage{colortbl}
\newcommand{\cmark}{\ding{51}}%
\newcommand{\xmark}{\ding{55}}%

\usepackage{hyperref}

\usepackage{microtype}

\aclfinalcopy 
\def\aclpaperid{***} 

\title{Reasoning Over History: Context Aware Visual Dialog}

\author{Muhammad A. Shah \thanks{$\quad$ equal contribution} \\
  \texttt{mshah1@cmu.edu} \\\And
  Shikib Mehri\footnotemark[1]  \\
  \texttt{amehri@cmu.edu} \\
  Language Technologies Institute\\
  Carnegie Mellon University\\
  Pittsburgh, PA\And
  Tejas Srinivasan\footnotemark[1] \\
  \texttt{tsriniva@cmu.edu}}

\date{}

\begin{document}
\maketitle
\begin{abstract}
While neural models have been shown to exhibit strong performance on single-turn visual question answering (VQA) tasks, extending VQA to a multi-turn, conversational setting remains a challenge. One way to address this challenge is to augment existing strong neural VQA models with the mechanisms that allow them to retain information from previous dialog turns. One strong VQA model is the MAC network, which decomposes a task into a series of attention-based reasoning steps. However, since the MAC network is designed for single-turn question answering, it is not capable of referring to past dialog turns. More specifically, it struggles with tasks that require reasoning over the dialog history, particularly coreference resolution. We extend the MAC network architecture with Context-aware Attention and Memory (CAM), which attends over control states in past dialog turns to determine the necessary reasoning operations for the current question. MAC nets with CAM achieve up to 98.25\% accuracy on the CLEVR-Dialog dataset, beating the existing state-of-the-art by 30\% (absolute). Our error analysis indicates that with CAM, the model's performance particularly improved on questions that required coreference resolution.
\end{abstract}

\section{Introduction}

Visual dialog is the task of answering a sequence of questions about a given image such that responding to any one question in the dialog requires context from the previous dialog history. The task of visual dialog \citep{das2017learning,kottur2019clevr} brings together several fundamental building blocks of intelligent systems: visual understanding, natural language understanding and complex reasoning. 
The multimodal nature of visual dialog requires approaches that jointly model and reason over both modalities. Furthermore, visual dialog necessitates the ability to resolve visual coreferences, which arise when two phrases in the dialog refer to the same object in the image. Visual coreference resolution requires both an ability to reason over coreferences in the dialog, as well as ground the entities from the language modality in the visual one.

In contrast to large-scale realistic datasets for visual dialog, such as VisDial \citep{das2017learning}, \citet{kottur2019clevr} introduce CLEVR-Dialog as a diagnostic dataset for visual dialog. In contrast to other visual dialog datasets, CLEVR-Dialog is synthetically generated - this allows it to be both large-scale and structured in nature. This diagnostic dataset allows for improved fine-grained analysis, using the structured nature of the images and language. This fine-grained analysis allows researchers to study the different components in isolation and identify bottlenecks in end-to-end systems for visual dialog.

Highly structured models have performed well on visual question answering and visual dialog \citep{andreas2016neural,andreas2016learning,kottur2018visual} by leveraging explicit \textit{program modules} to perform composition reasoning. CorefNMN \citet{kottur2018visual}, which leverages explicit program modules for coreference resolution, was the previous state-of-the-art model on the CLEVR-Dialog dataset. However, the explicit definition of \textit{program modules} requires handcrafting and limits generalizability. As such, we explore mechanisms of relaxing the structural constrains by using MAC Networks \citep{hudson2018compositional} and adapting it to the task of dialog. Specifically, we introduce the Context-aware Attention and Memory (CAM) to serve as an inductive bias that allows MAC networks to explicitly capture the necessary context from the dialog history.

CAM consists of a context-aware attention mechanism and a multi-turn memory state. The context-aware attention mechanism attends over the control states of past dialog turns, to determine the control states for the current dialog turn. Since control states in a MAC networks are analogous to program modules, the attention effectively leverages past reasoning operations to inform current reasoning operations. For example, if the MAC network had to locate the \textit{``the red ball''}, a future turn which refers to \textit{``the object to the left of the previous red object''} can attend to the control state responsible for locating the red ball. Meanwhile the multi-turn memory \textit{remembers} information extracted to answer previous questions in the dialog. Similar to the explicit programs of CorefNMN, CAM serves to model properties of dialog (e.g., coreference resolution, history dependent reasoning). However, unlike CorefNMN, CAM does not require explicit handcrafting, can be trained end-to-end and is capable of generalization.

Our methods attain state-of-the-art performance on CLEVR-Dialog, with a \textbf{30\% improvement} over the prior work. Further, CAM provides strong performance gains over MAC networks across several different experimental setups. Analysis shows that CAM's attention weights are meaningful, and particularly useful for questions that require coreference resolution across dialog turns.

\section{Related Work}

\subsection{Visual Question Answering}

Visual Question Answering (VQA) requires models to reason about an image, conditioned on a complex natural language question. Solving VQA requires the ability to reason over images and grounding language entities in the visual modality. There have been several datasets proposed for this task, such as the open-ended VQA \citep{antol2015vqa} and diagnostic CLEVR \citep{johnson2017clevr} datasets, and several models proposed to solve this task \citep{yu2015visual, malinowski2014multi, gao2015you, ren2015exploring, liu2019clevr}.

State-of-the-art modeling approaches to VQA can largely be broken into two categories: modular networks \citep{yi2018neural, andreas2016neural, hu2017learning}, and end-to-end differentiable networks \citep{hudson2018compositional}. Neural Module Networks (NMNs) consist of specialized neural modules and can be composed into programs. Since the program construction is not differentiable, training module networks involves complex reinforcement learning training techniques. Moreover, the strong structural constraints along with the need to handcraft modules limits the generalizability of these models. We think that relaxing some structural constraints, such as those involved in handcrafted models, while retaining other, specifically those that allow for compositional reasoning, would yield powerful yet flexible models.

As a step in this direction, \citet{hudson2018compositional} have proposed MAC Networks (Memory, Attention and Comprehension Networks) which simulates a $p$-step compositional reasoning process by decomposing the question into a series of attention-based reasoning steps. Unlike NMNs, MAC networks do not have specialized program modules, instead they use a control unit to predict a continuous valued vector representation of the reasoning process to be performed at each step.

\subsection{Visual Dialog}

As models achieve human-level performance on VQA, \citet{das2017visual} and \citet{kottur2019clevr} proposed to extend them to a conversational setting. Concretely, visual dialog is a multi-turn conversation grounded in an image. In addition to the challenges of VQA, visual dialog requires reasoning over multiple turns of dialog, in which can refer to information introduced in previous dialog turns.

Several datasets have been introduced to study the problem of visual dialog, such as the large-scale VisDial dataset \citep{das2017visual} and the diagnostic CLEVR-Dialog dataset \citep{kottur2019clevr}. CLEVR-Dialog is a programmatically constructed dataset with complex images and conversations reasoning about the objects in a given image. Similar to the CLEVR dataset, CLEVR-Dialog comprises of queries and responses about entities in a static image. However, in this multi-turn dataset, queries make references to entities mentioned in previous turns of the dialog, and can thus not be treated as single-turn queries. The main challenge in CLEVR-Dialog is thus visual coreference resolution - resolving multiple references across dialog turns to the same entity in the image. 

Several recently proposed methods use reinforcement learning techniques to solve this problem \citep{strub2017end, das2017learning}. \citet{strub2017end} policy gradient based method for visually grounded task-oriented dialogues. On the other hand, \citet{das2017learning} utilise a goal-driven training for visual question answernig and dialog agents via a cooperative game between two agents (questioner and answerer) and learn the policies of these agents using deep reinforcement learning.

Other approaches utilise transferring knowledge from a discriminatively trained model to a generative dialog model \citep{lu2016hierarchical} and by using differentiable memory to solve visual coreferences \citep{seo2017visual}. More specifically, \citet{seo2017visual} utilise an associative attention memory for retrieving previous attentions which are most useful for answering the current question. Later, the retrieved attention is combined with a tentative one via dynamic parameter prediction in order to answer the current question.

\citet{kottur2018visual} adapted NMNs used in \citep{andreas2016neural} with an addition of two modules (Refer and Exclude) specifically meant for handling coreference resolution. These two modules perform explicit coreference resoltuion at a word level granularity. The module 'Refer' grounds coreferences in the conversation history while 'Exclude' handles contextual shifts. This Coreference Neural Module Networks (Coref-NMN) were applied to CLEVR-Dialog \citep{kottur2019clevr}, and achieve the best accuracy on the dataset.

\section{Methods}
\subsection{Problem Statement}
Formally, the task we are tackling in this paper is to pick the correct answer, $a^*_t\in\mathcal{A}$ for a question, with representation $q_t\in\mathcal{Q}$, based on an image, $I\in\mathcal{I}$, with a caption, $C\in\mathcal{C}$, and a past dialog history, $\mathcal{H}_t=\{(q_1, a_1), (q_2, a_2) .. (q_{t-1}, a_{t-1})\}$, where $a_i$ is the answer to question $q_i$. In practice, $\mathcal{I}$ does not contain the actual image, but rather, an embedding of the image computed using a pretrained image recognition model.

\subsection{MAC Network Architecture}
Since our approach builds upon the MAC Network architecture \cite{hudson2018compositional}, we will briefly introduce it in this section before presenting our own novel extensions to it in the subsequent sections. 

\begin{figure}
    \centering
    \includegraphics[width=0.9\linewidth]{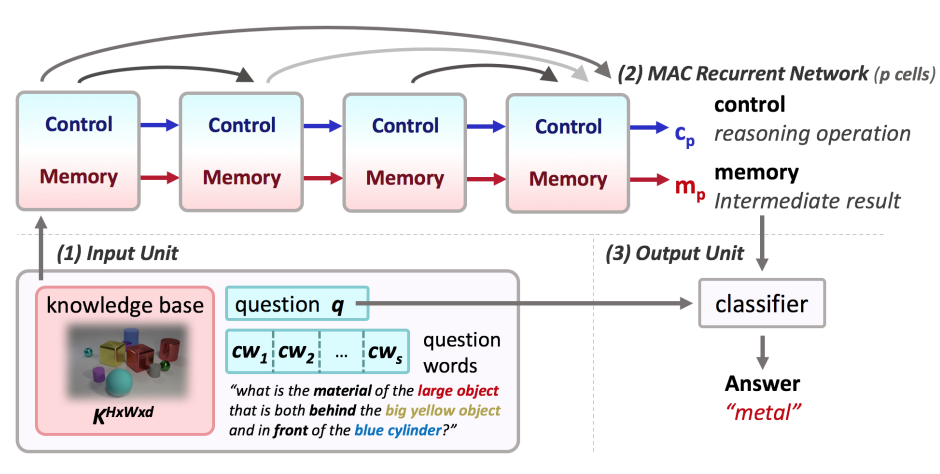}
    \caption{The MAC Network Architecture (image from \cite{hudson2018compositional})}
    \label{fig:macnet}
\end{figure}

The MAC network has three core components: the input unit, the MAC cell and the output unit. The input unit computes an image representation, a single question embedding for the entire question and contextualized word embeddings for each word in the question. The output of the input unit is recurrently passed through the MAC cell $p$ times, where $p$ is a predefined hyper-parameter. Each pass through the MAC cell is meant to simulate one step of a $p$-step reasoning process. The MAC cell consists of three sub-modules, namely the control, read and write units and a running memory state that accumulates the results of each reasoning step. The control unit computes an embedding for the $i^{\text{th}}$ reasoning operation based on the $(i-1)^{\text{th}}$ reasoning operation, and the sentence and word embeddings of the question. The read unit attends on the image representation using the current memory state and the output of the control unit, to extract the information required for current reasoning step. The write unit uses the output of the read unit to update the memory state. After $p$ reasoning steps have been performed, the output unit uses the memory state to predict the answer. It is assumed in the architecture that the answers are categorical. This process is illustrated in Figure \ref{fig:macnet}.

\subsection{Extending MAC Network With CAM}
Since the MAC network is designed to answer single-turn questions, it is not able to answer questions that rely on context established in previous turns of a dialog. In this section we describe our proposed Context-aware Attention and Memory (CAM) mechanism that endows MAC networks with the ability to perform multi-turn reasoning by \textit{remembering} the reasoning steps it performed, and the information it extracted from the image to answer questions posed in past turns. CAM has two components, namely a memory state that remains persistent across multiple turns and an attention mechanism that encodes contextual information from past turns in the current control state. 

\subsubsection{Multi-Turn Memory}
The first extension we propose endows the model with a memory that remains persistent across dialog turns. Specifically, we want to allow the model to \textit{remember} the information it has already extracted from the image in earlier dialog turns, so that it can use this information to answer context-dependent questions in subsequent turns.

To implement this memory mechanism we leverage the existing memory state of MAC networks, with a slight modification. In the original MAC network architecture the memory state is initialized with a zero vector for each question, and is updated after each of the $p$ reasoning steps before being discarded. Formally, the memory state at the $k^{\text{th}}$ reasoning step of the $t^{\text{th}}$ turn in the dialog is computed as
\begin{equation}
    m^{(t)}_k = 
    \begin{cases}
    f(I^{(t)}_{k}, \mathbf{0}),& k = 0\\
    f(I^{(t)}_{k}, m^{(t)}_{k-1}), & k > 0
\end{cases}
\label{eq:mem-update}
\end{equation}
where $I^{(t)}_{k}$ represents the information extracted from the image and $f$ is a function that computes the updated memory state.
Under this scheme, the information accumulated while reasoning about the first question in the dialog is discarded when the model starts reasoning about the second question. 
In our implementation we initialize the memory once for each dialog, and retain it across all the turns of the dialog. Formally, this leads to the modification of Equation \ref{eq:mem-update} to
\begin{equation}
    m^{(t)}_k = 
    \begin{cases}
    f(I^{(t)}_{k}, \mathbf{0}),& t=0,k = 0\\
    f(I^{(t)}_{k}, m^{(t-1)}_{k}),& t>0,k = 0\\
    f(I^{(t)}_{k}, m^{(t)}_{k-1}), & k > 0
\end{cases}
\end{equation}

\subsubsection{Context-aware Attention Mechanism}
For our second extension, we propose to allow the model to \textit{recall} previous control states when computing the current control state. The intuition behind this extension is that if the current question, $q_t$, references an entity from a previous question, $q_{t-k}$ or its answer, the reasoning steps for answering $q_t$ are likely to be similar, to those for answering $q_{t-k}$, at least insofar as they relate to the coreferent entity. Since the introduction of the coreferent entity was more recent with respect to $q_{t-k}$, compared to $q_t$, it would have been more salient in the model's memory. Therefore, it is likely that the model would have applied appropriate reasoning processes in when answering $q_{t-k}$. At $q_t$, the coreferent entity is less salient in the model's memory, which increases the likelihood of the model selecting the inappropriate reasoning steps. 

\begin{figure}
    \centering
    \includegraphics[width=0.40\textwidth]{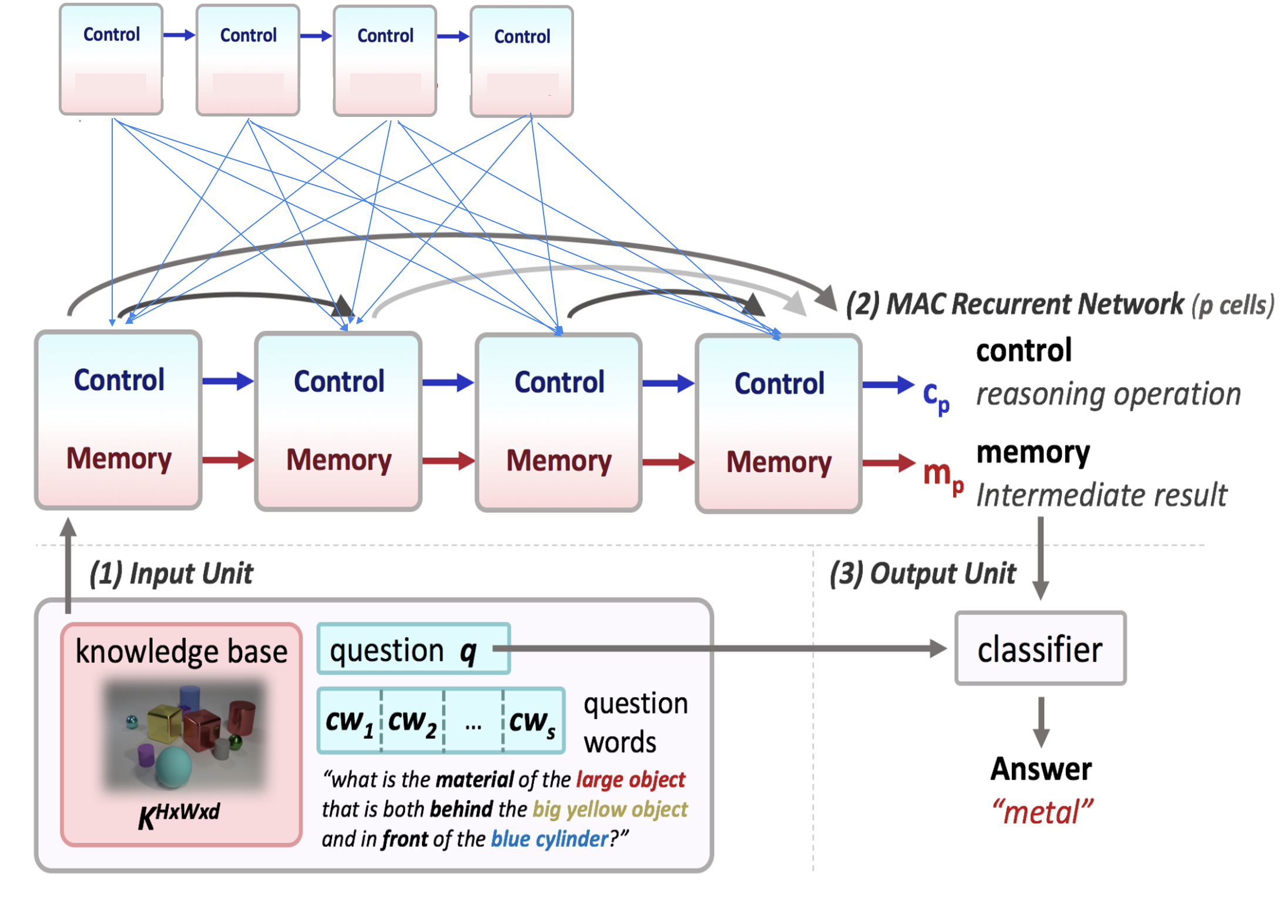}
    \caption{The modified MAC Network architecture which explicitly incorporates dialog context}
    \label{fig:macnet-att}
\end{figure}

To mitigate the aforementioned problem and explicitly incorporate the dialog context into the model, we introduce a transformer-like self-attention mechanism on the previous control states. The resulting architecture is illustrated in Figure \ref{fig:macnet-att}. This mechanism allow the model to \textit{explicitly} attend to the past outputs of the control unit, both, from previous reasoning steps in the current turn and the reasoning steps from the previous dialog turns, while computing the control output for the current reasoning step. 

Concretely, given the unattended control representations, ${\mathbf{C}=[c_1^{(1)} ... c_{i-1}^{(t)}]^T}$ of all the reasoning steps until the $i^{\text{th}}$ reasoning step of turn $t$, the final control output, $\Bar{\mathbf{C}}$, is computed as the \textit{fusion} \cite{hu2017learning} of the attended control representation, $\Hat{\mathbf{C}}$, and the unattended control representation as follows:
\begin{align}
    E &= \phi_{key}(\mathbf{C})\phi_{key}(\mathbf{C})^T\\
    A &= \text{softmax}(\text{tril}(E))\\
    \Bar{\mathbf{C}}&=fusion(\mathbf{C}, A\mathbf{C})
\end{align}
where $\phi_{key}$ and $\phi_{value}$ represent the key and value projections used in the self-attention step, $\text{tril}(E)$ represents the matrix obtained by setting the values in the upper triangle of $E$ to zero and the \textit{fusion} module is defined as follows:

\begin{align}
    \text{fusion}(x,y) = g\Tilde{x} + (1-g) x\nonumber\\
    \Tilde{x} = relu(W_r[x];y;x\odot y; x-y)\nonumber\\
    g = sigmoid(W_g[x];y;x\odot y; x-y) \nonumber
\end{align}

where $\odot$ represents element-wise multiplication.

\section{Experimental Setup}

\subsection{Dataset}

\begin{figure}
    \centering
    \includegraphics[width=0.25\textwidth]{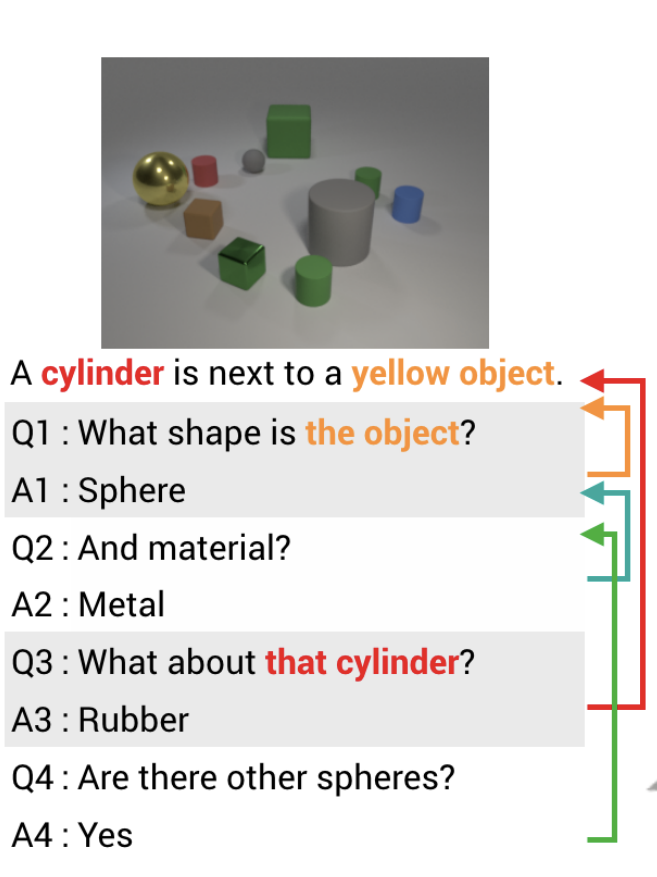}
    \caption{Example from the CLEVR-Dialog dataset, consisting of an image, a dialog. Each dialog begins with a caption describing the image, followed by a round of questions and answers. Each question relies on information from previous dialog turns.}
    \label{fig:clevr_dataset}
\end{figure}

The CLEVR-Dialog dataset\footnote{\url{https://github.com/satwikkottur/clevr-dialog}} \citep{kottur2019clevr}, pictured in Figure \ref{fig:clevr_dataset}, consists of several modalities: visual images, natural language dialog and structured scene graphs. 

Each image $I$ and its respective complete scene graph $S_a$ depicts a scene containing several objects. Each object has four major attributes, enumerated as follows:

\begin{itemize}
    \itemsep0em
    \item Color -- \textit{blue, brown, cyan, gray, green, purple, red, yellow}
    \item Shape -- \textit{cylinder, cube, sphere}
    \item Size -- \textit{large, small}
    \item Material -- \textit{metal, rubber}
\end{itemize}

Every pair of objects has a spatial relationship which describes their relative spatial position: \textit{front, back, right, left}.

Each dialog is an interaction between a Questioner and an Answerer. The Answerer, who has the image and the complete scene graph, begins by providing a caption that describes the image. The Questioner, who does not see the image, aims to build up a complete scene graph by repeatedly asking questions. As the Questioner gets more information, they build up a partial scene graph $S_q^t$. Though, during data collection the Answerer had a complete scene graph, during the task of visual dialog -- the scene graph should not be used during testing and the Answerer can only use the image. 

Each dialog consists of 10 turns. Questions are generated with the use of 23 question templates, which can be grouped into several categories: \textbf{Count} questions which ask for the number of objects that satisfy certain conditions, \textbf{Existence} questions are yes/no questions that query about certain conditions in the image and \textbf{Seek} questions ask for attributes of certain objects. The seek question type is 60\% of the dataset, followed by count at 23\% and exist at 17\%. There are 29 unique answers (e.g., 'yes', 'no', 'blue', '1', '2' etc.), with all the answers being single-word.

A strong motivation of the CLEVR-Dialog dataset is to model \textit{dialog} in the context of an image. To this end, there are two types of history dependancy. The first is coreference, wherein a phrase in a question refers to an earlier referent in the history. The mean coreference distance is 3.2 turns and the distance spans between 1-10 turns. The second type of history dependency is when the question relies on the entire dialog history, rather than a specific referent. For example: \textit{'How many other objects are there?'}

The dataset has 85\textit{k} unique images, with 5 dialogs per image for a total of 425\textit{k} dialogs. Each dialog consists of a caption, and ten turns of question-answer pairs, for a total of 4.25\textit{M} questions and answers. There are 23 unique question templates and 73\textit{k} unique questions. 

Since CLEVR-dialog has 29 unique single-word answers, the metric used is accuracy. The structured nature of the dataset allows for accuracy breakdown by coreference distance and question type, as shown by \citet{kottur2019clevr}.
 
\subsection{Implementation Details}

All models were implemented in PyTorch \footnote{\url{https://github.com/tohinz/pytorch-mac-network}} building on an open source implementation of MAC networks. While training history agnostic models each dialog turn was treated as an independent question and in each iteration we trained the model on 128 random dialog turns. Meanwhile, we trained the context-aware models by providing them one dialog turn at a time, with a batch size of 12 dialogs (120 turns). The learning rate and the number of reasoning steps for the MAC networks was set to \texttt{2e-4} and 8, respectively. 

Since CLEVR-Dialog consists of only a training set and a development set, the development set was used for evaluation. We remove 1000 images and their respective dialogs from the training set to use for validation, for a total of 5000 dialogs and 50,000 dialog turns. The models were set to trained for 25 epochs but if the validation accuracy does not increase for 5 epochs, we stop training so some models were trained for 16-17 epochs while others were trained for 25. We ran experiments on a cluster with 32 core Intel Xenon processors and Nvidia 1080Ti GPUs.

\begin{table}
\centering\small
\begin{tabular}{|l|c|c|c|c|}
\hline
Model                         & CQ               & CAA & MTM      & Accuracy (\%) \\ \hline \hline
NMN     &   - &     -   & - & 56.6 \\ \hline 
CorefNMN                      & -                     & -                       & -                     & 68.0     \\ \hline \hline
\multirow{6}{*}{MAC Networks} & \xmark & \xmark   & \xmark & $65.9$   \\ \cline{2-5} 
                            & \xmark & \cmark   & \xmark & $89.43^*$ \\
\cline{2-5}
                              & \xmark & \cmark   & \cmark & $97.98^*$    \\ \cline{2-5} 
                              & \cmark & \xmark   & \xmark & $98.08^*$    \\ \cline{2-5} 
                              & \cmark & \cmark   & \cmark & $98.16^\dagger$      \\ \cline{2-5}
                              & \cmark & \cmark   & \xmark & $98.25^*$    \\ \hline 
\end{tabular}
\caption{Performance of our baseline models, and the effect of our dialog-specific augmentations, namely Concatetating Questions (CQ), Context-Aware Attention (CAA) and Multi-Turn Memory (MTM), to the MAC network \citep{hudson2018compositional}.\\ 
$^* p<0.00001$ compared with previous row.\\
$^\dagger p<0.001$ compared with previous row.
}
\label{tab:results}
\end{table}

\subsection{Baselines}

We use Neural Module Networks \cite{andreas2016neural} (NMN) and CorefNMN \cite{kottur2018visual} as our baslines because the dataset paper \cite{kottur2019clevr} reports them to have the best performance on CLEVR dialog.

Neural Module Networks (NMN) proposed by \citet{andreas2016neural} are a general class of recursive neural networks \citep{socher2013recursive} which provide a framework for constructing deep networks with dynamic computational structure. NMNs are history agnostic, making them a weak baseline for this dataset.

CorefNMNs \citep{kottur2018visual} adapts NMNs \citep{andreas2016neural} with an addition of two modules (Refer and Exclude) that perform explicit coreference resolution at a word level granularity. `Refer' grounds coreferences in the conversation history while `Exclude' handles contextual shifts.  

\section{Results}

In Table \ref{tab:results}, we examine the performance of our baseline models, and the effect of Context-aware Attention Mechanism (CAM). We experiment with three different combinations of our dialog-specific extensions to the MAC network architecture:
\begin{enumerate}[label=(\roman*)]
    \itemsep-0.5em
    \item context-aware attention over control states,
    \item multi-turn memory, and
    \item concatenating the dialog history as input to MAC - an obvious but naive and inefficient strategy for incorporating contextual information into a single-turn QA model.
\end{enumerate}

When none of these three extensions are present, we obtain the vanilla MAC network which does not have the ability to reason over the dialog context.  We see that vanilla MAC achieves 10\% higher accuracy than NMN, which is also history-agnostic, and is surprisingly close to CorefNMN, which explicitly reasons over the dialog history. The fact that a single turn model can correctly answer two-thirds of the questions in a very large dataset raises some concerns regarding how representative is the dataset of an actual dialog task.
\begin{figure*}
    \centering
    \includegraphics[width=0.85\textwidth]{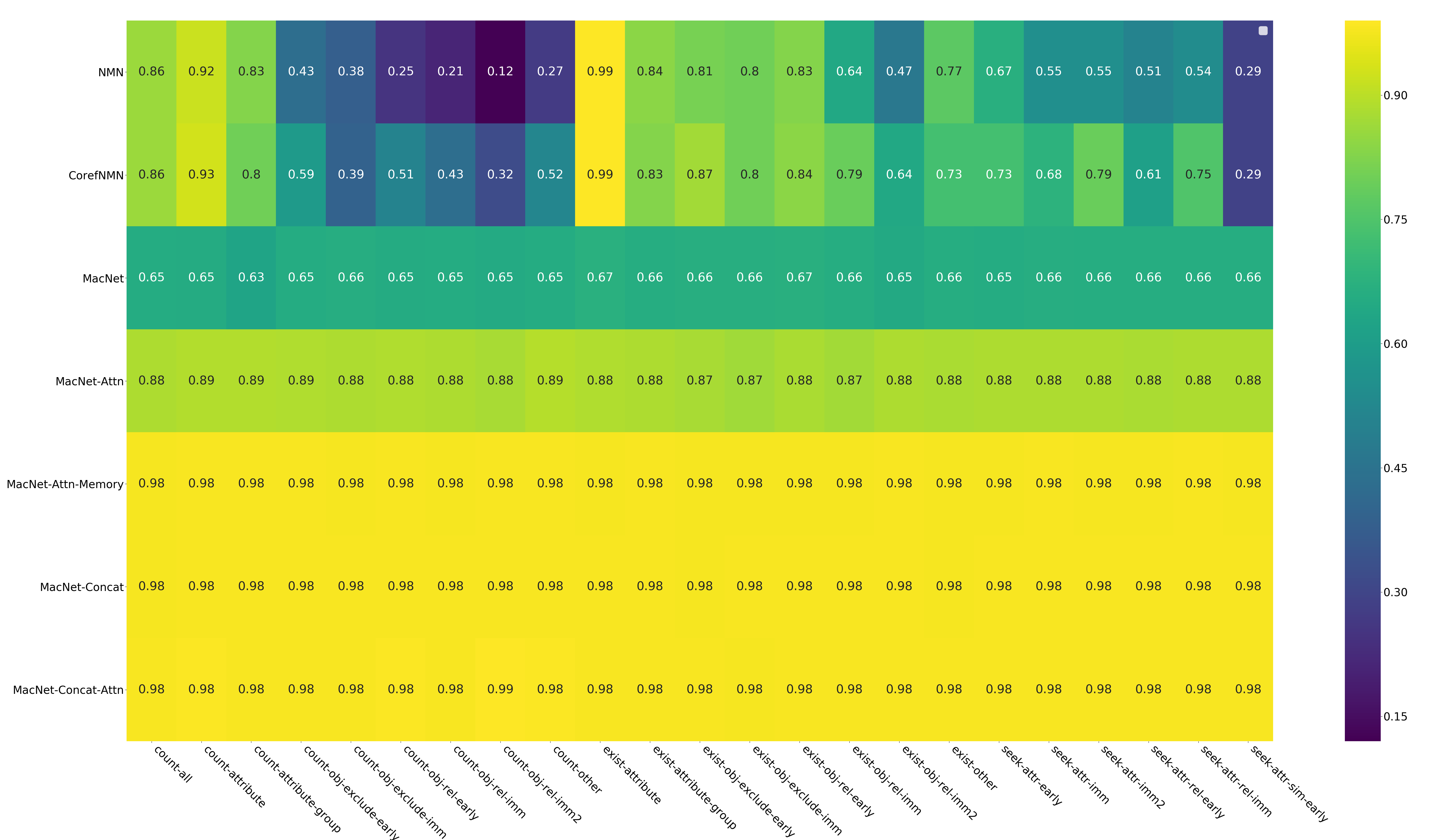}
    \caption{Breakdown of the accuracies of the models by question type. Different question types require different reasoning, especially pertaining to the dialog history.}
    \label{fig:by_question_type}
\end{figure*}
Adding context-aware attention to the MAC improves the accuracy of the model considerably to 89.43\%. Introducing multi-turn memory to this model yields an accuracy of 97.98\% accuracy - an improvement of 30\% (absolute) on the performance of vanilla MAC network and benchmark CorefNMN. These results emphatically establish the efficacy of CAM and establish a new state-of-the-art for CLEVR-Dialog.

Perhaps most notably, concatenating the dialog history to the current query works remarkably well for MAC networks, achieving 98.08\% accuracy with no other augmentations to the MAC network. Introducing context-aware attention further improves accuracy to 98.25\%, which yet again evidences the efficacy of the attention mechanism we propose. However, introducing the multi-turn memory results in a slight decrease in performance, indicating that the memory mechanism is not useful when the entire dialog context is present. 

We think it is important to mention here that, concatenating the dialog history is a naive method, and this method becomes computationally inefficient when the dialog history is longer and the questions themselves are longer. While it is true that the context aware attention mechanism also stores additional data - the past control states. However, since the control states are fixed sized, the additional memory and computation required is in $O(Tp)$, where $T$ is the maximum number of dialog turns and $p$ is the number of reasoning steps to be performed. On the other hand, the increase memory and computation requirements for concatenation is in $O(|Q_{max}|Tp)$ where $|Q_{max}|$ is the length of the longest question. Without concatenating the dialog history, incorporating the multi-turn memory greatly improves accuracy (89.43\% $\rightarrow$ 97.98\%), while being more computationally efficient.

These results indicate that relaxing the structure of the model by eliminating hand-crafted modules gives the model much more flexibility in how it processes the input query, allowing it to perform more complex reasoning than the programs assembled by Neural Module Networks.

\begin{figure}[h]
    \centering
    \includegraphics[width=0.9\columnwidth]{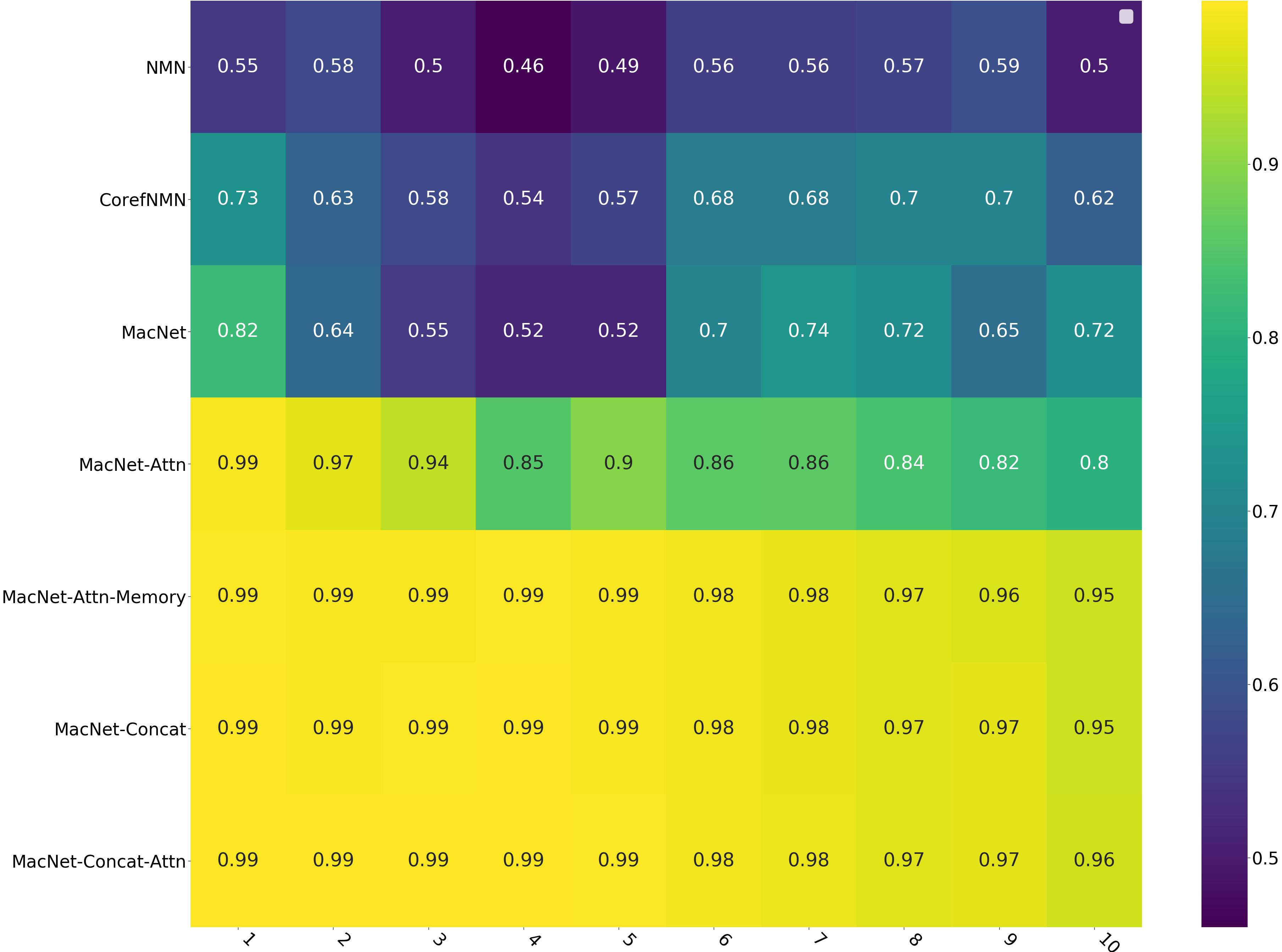}
    \caption{Breakdown of the accuracies of different models on different turns in the dialog}
    \label{fig:by_hist}
\end{figure}

\section{Analysis}

\subsection{Performance Breakdown}
\begin{table*}
\centering
\begin{tabular}{l|l}
\hline
                                 & \cellcolor[HTML]{FFC4C4}\small  There are 4 small things. What is the number of green things in the view, if present? \\
                                 & \cellcolor[HTML]{FFC9C9}\small Are there other things that share its color in the scene?                                                   \\
\multirow{-3}{*}{\small Previous Turns} & \cellcolor[HTML]{FFADAD}\small If there is a thing in front of the above green thing, what is its material?                                \\ \hline
\small Current Turn                     & \cellcolor[HTML]{FFC2C2}\small If there is a thing to the right of it, what color is it?                                                   \\ \hline \hline
                                 & \cellcolor[HTML]{8F8FFF}\small The image has a yellow thing right of a cylinder. How many other things are in the picture?                  \\
                                 & \cellcolor[HTML]{D4D4FF}\small What is the size of the previous cylinder?                                                                  \\
\multirow{-3}{*}{\small Previous Turns} & \cellcolor[HTML]{D4D4FF}\small Does the previous yellow thing have things to its behind?                                                   \\ \hline
\small Current Turn                     & \cellcolor[HTML]{D4D4FF}\small If there is a thing behind the previous cylinder, what is its material?                  \\ \hline
\end{tabular}
\caption{Dialog examples where attention over control states of previous dialog turns informs the model of which previous turn is important to attend to when answering the current query (the last question of the dialog). Darker shade means higher attention weight.}
\label{tab:attn-examples}
\end{table*}
In order to better understand the results presented above, we breakdown the accuracy of the models along dialog turns and question types. 
\subsubsection{Question Types and Accuracy}
The heatmap shown in Figure \ref{fig:by_question_type} breaks down the accuracies of the models for different question types. Different question types require different reasoning about the image and the dialog. For example, \texttt{count-obj-rel-*} requires the models to count the number of objects \textit{relative} to another entity, often one that was discussed earlier in the dialog. We observe that MacNet-Concat-Attn obtains a 1\% gain over MacNet-Concat and MacNet-Attn-Memory for the \texttt{count-obj-rel-imm2} question type, which requires reasoning about the number of objects relative to one from earlier in the dialog. These question types are follow-ups (e.g., ``\textit{how about to it's left}''), meaning that they have both anaphora and ellipsis. As such the performance gains on this question type are indicative of better dialog modelling. It is important to note that the above question types are also the ones that have the lowest performance across all models. This highlights the importance of developing specialized strategies for modelling dialog.
\subsubsection{Dialog Turn Number and Accuracy}
The heatmap shown in Figure \ref{fig:by_hist} presents the accuracy of the models when answering questions at different dialog turns. MacNet-Attn performs significantly better than MacNet. The fact that MacNet-Attn performs better at later turns suggests that the model is effectively resolving coreferences from the dialog history. Likewise, MacNet-Attn-Memory obtains even stronger performance gains, especially at later dialog turns. In the final turn of dialogs, MacNet-Attn-Memory is 15\% more accurate than MacNet-Attn and 23\% over MacNet.

MacNet-Concat-Attn obtains a 1\% improvement over MacNet-Attn-Memory and MacNet-Concat, at the $9^{th}$ and $10^{th}$ dialog turns, respectively. This performance gain is relatively smaller, however, since the accuracies are so high, the relative error reduction is still significant. It is important to note that a 1\% improvement in accuracy corresponds to answering 7500 more questions correctly. 

\subsection{Attention Analysis}
We verify that the context-aware attention over the control states is performing coreference resolution by looking at the attention weights assigned to each past question in the dialog history. Since 8 control states are computed per question, we consider the maximum attention weight between any control state of the current question and any control state in the past question.

Table \ref{tab:attn-examples} presents examples of turn-level attention weights for two different dialogs (in red and blue, respectively). The first example shows that a higher attention weight is allotted to the immediately preceding dialog turn. We note that this preceding turn contains a reference to the entity which is referred to in the current turn.

In the second example, we see that a much higher attention is given to the first turn (which includes the image caption). We noticed that lots of dialog turns give a higher attention to the first dialog turn. This could be because a lot of questions start a new line of dialog by making a reference back to the original image caption. For instance, in the current turn, a question is asked about an entity in relation to the cylinder which is mentioned in the caption.

These examples illustrate that CAM is able to identify the referent turn in the dialog and appropriately attend to it.

\section{Conclusion}

We present Context-aware Attention and Memory (CAM), a set of dialog-specific augmentations to MAC networks \citep{hudson2018compositional}. CAM consists of a context-aware attention mechanism which attends over the MAC control states of past dialog turns and a persistent, multi-turn memory which is accumulated over multiple turns of the dialog. These augmentations serve as an inductive bias that allow the architecture to capture various important properties of dialog, such as coreference and history dependency. 

Our methods attain state-of-the-art performance on CLEVR-Dialog, with our best model attaining an accuracy of \textbf{98.25\%}, a \textbf{30\% improvement} over all prior results. Further, CAM attains strong performance gains over vanilla MAC networks, especially for question types that require coreference resolution and later dialog turns. Ablation experiments indicate that both of the components in CAM provide significant improvement in performance. We also verified that the context aware attention mechanism in indeed captures coreferences between the dialog turns.

Our results are indicative of the flexibility of weakly structured models, like MAC networks and their ability to adapt to different problem settings. To adapt MAC networks for visual dialog we had to devise a mechanism to provide it with contextual information from past turns. Thereafter, the other components of the model were able to adapt and use this information to improve performance on the task, whereas to adapt NMN to visual dialog \citet{kottur2018visual} had to devise specialized modules to handle specific types of questions.

\bibliographystyle{acl_natbib}
\bibliography{icml}

\end{document}